\documentclass[10pt,conference]{IEEEtran}
\usepackage{cite}
\usepackage{amsmath,amssymb,amsfonts}
\usepackage{algorithmic}
\usepackage{graphicx}
\usepackage{textcomp}
\usepackage{xcolor}
\def\BibTeX{{\rm B\kern-.05em{\sc i\kern-.025em b}\kern-.08em
    T\kern-.1667em\lower.7ex\hbox{E}\kern-.125emX}}
\usepackage{cleveref}
\usepackage{listings}
\usepackage{wrapfig}
\usepackage[para]{footmisc}

\begin{document}

\title{Reproducibility Requires Consolidated Artifacts}

\author{\IEEEauthorblockN{Iordanis Fostiropoulos}

\and
\IEEEauthorblockN{Bowman Brown}
\IEEEauthorblockA{University of Southern California, Los Angeles, CA\\
\texttt{\{fostirop,bnbrown,itti\}@usc.edu}
}
\and
\IEEEauthorblockN{Laurent Itti}
}
\maketitle

\begin{abstract}
Machine learning is facing a `reproducibility crisis' where a significant number of works report failures when attempting to reproduce previously published results. We evaluate the sources of reproducibility failures using a meta-analysis of 142 replication studies from ReScience C and 204 code repositories. We find that missing experiment details such as hyperparameters are potential causes of unreproducibility. We experimentally show the bias of different hyperparameter selection strategies and conclude that consolidated artifacts with a unified framework can help support reproducibility.
\end{abstract}

\section{Introduction}
\label{sec:intro}

Evaluating ML research requires reproduction studies such as those in ReScience C \footnote{http://rescience.github.io/}. However, reproduction studies cost researchers time, energy, and resources and can call into question the validity of the experiment being analyzed.

\cite{natureTransparentAI} identifies missing and convoluted artifacts as one of the main causes of non-reproducible research. Artifacts include configuration details, details on the methodology, and code. Additionally, results can be sensitive to both the hyperparameter selection strategy and the computational budget, and differences can lead to unreproducible results \cite{Shwartz-Ziv2021}. Current practices specify all of the hyperparameter details in different artifacts, i.e., within the paper text, hard-coded in the open-source code repository, or as default runtime arguments. Lastly, inter-dependencies between multiple software frameworks used for a single experiment can lead to unreproducibility \cite{mlDependencyUsageLibraryUsage}.

\begin{figure}
\centerline{\includegraphics[width=\linewidth]{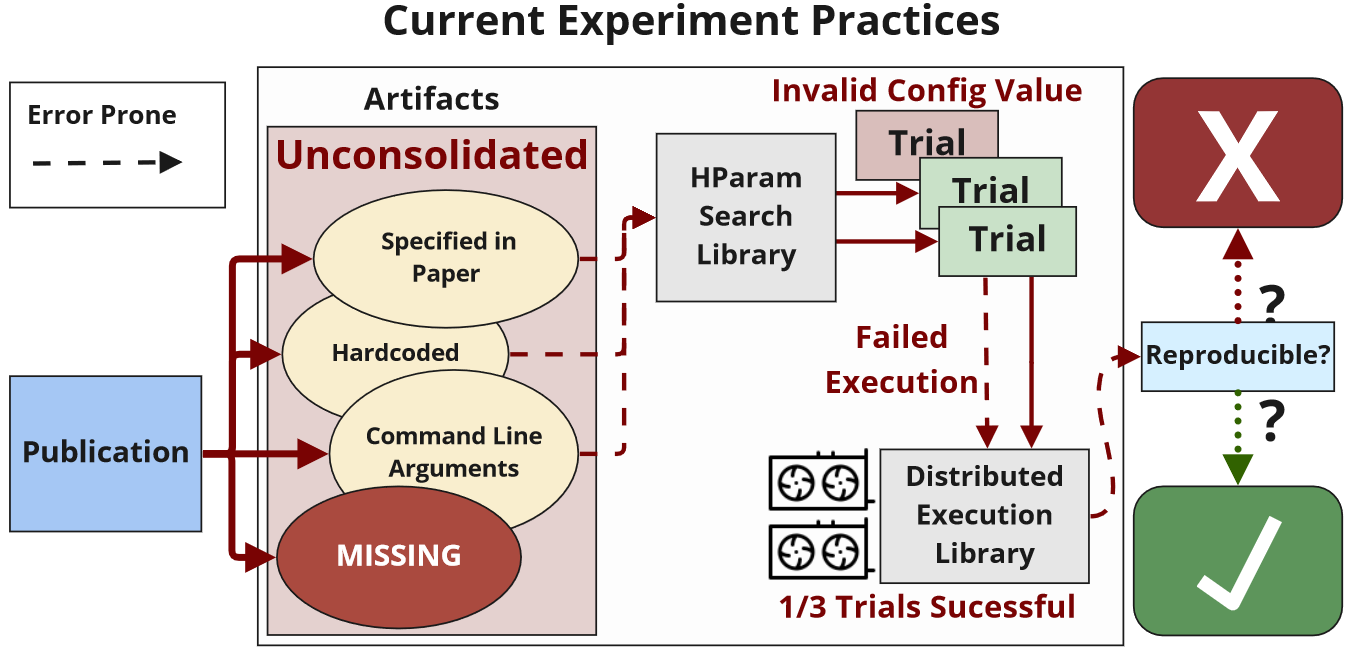}}
\caption{Hyperparameter values can be scatter; at each experiment step independent errors from the frameworks used can lead to failures, such as with \textit{inter-dependencies}. The end result is that only a subset of the original trials are valid, which can lead to biased analysis and unreproducible results.
}
\label{fig:sampling_strategy}
\end{figure}

We identify that additional effort by the replication studies lead to unreproducible outcomes caused by missing artifacts, 
complications arising from inter-project dependency errors between ML tools,  
and differences in hyperparameter selection strategies that lead to erroneous analysis.

\section{Reproducibility Analysis}
\label{sec:methods}

\subsection{Missing Artifacts}

We manually evaluate 142 papers that reproduce previous studies and are published at the open-access peer-reviewed journal ReScience C. We tagged each paper based on the issues the authors faced when reproducing the original work with tags denoting implementation issues, hyperparameter issues, and the responsiveness of the original author. 

\begin{wrapfigure}{L}{.5\linewidth}
\centerline{\includegraphics[width=\linewidth]{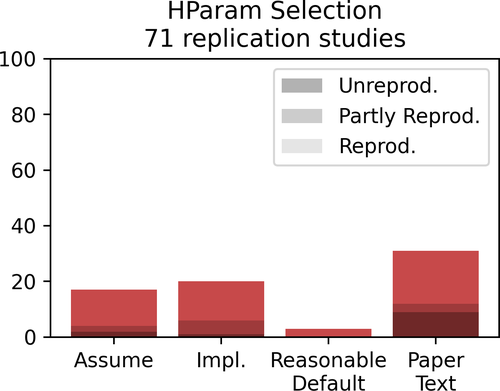}}
\caption{We identify the source of the hyperparameter for the replication studies for 71 of the 142 original studies for which the hyperparameters were unconsolidated. Hyperparameters scattered in the original study's text and artifacts led to unreproducible outcomes.}
\label{fig:hparam}
\end{wrapfigure}

Of the studies evaluated,  80.99\% were able to reproduce, 9.86\% were unable to reproduce, and 9.15\% were partially able to reproduce the results from the original study. %
Compared to the reproducible studies, unreproducibile studies were missing code 15.79\% more often and hyperparameter details 19.91\% more often.

The results of our survey suggest that the availability of the code, hyperparameter details and medium by which such details are shared were the most important factors for reproducibility.

\subsection{Problematic Tooling}

We quantitatively evaluate the inter-project dependency issue for reproducibility \cite{mlDependencyUsageLibraryUsage}. We mine 132 repositories from the 142 reproduction studies and 72 repositories corresponding to the official implementations from the original studies.

We consider a repository that uses any combinations of `Ray', %
`HyperOpt', %
`PyTorch-Ignite', %
`Optuna', %
`Hydra', %
and `PyTorch-Lighting', %
as using `multiple frameworks'. We identify repositories that include their hyperparameters in a file format (such as \texttt{.yaml}) as using a `consolidated configuration'.

Using \textit{multiple frameworks} has a Pearson correlation coefficient $r=-0.20$ with reproducible outcomes and  $r=0.24$ for partly unreproducible outcomes. In contrast, \textit{consolidated configuration} was correlated with reproducibility with $r=0.15$. We observe that defining hyperparameter values in multiple sources rather than in a consolidated location is error-prone. We found that when replication studies sourced the hyperparameter details scatter in the paper-text were proportionally more likely to have unreproducible results compared to sourcing them from the code \cref{fig:hparam}.

\subsection{Invalid Tuning Comparisons}

To evaluate the bias introduced by different hyperparameter search strategies during ML experimentation, we use Emukit\cite{emukit} to sample hyperparameters from regions of high accuracy (`Greedy'), both high accuracy and high variance (`CombinedVariance'), and high variance alone (`ModelVariance'). We use Optuna \cite{Optuna} to evaluate a Tree-structured Parzen Estimator (`TPE'). We also test quasi-random `Sobol' sequence sampling and pseudo-random (`Random') sampling.

\begin{figure}%

\begin{minipage}[t]{.7\linewidth}
  \centering\raisebox{\dimexpr \topskip-\height}{%

\centerline{\includegraphics[width=\linewidth]{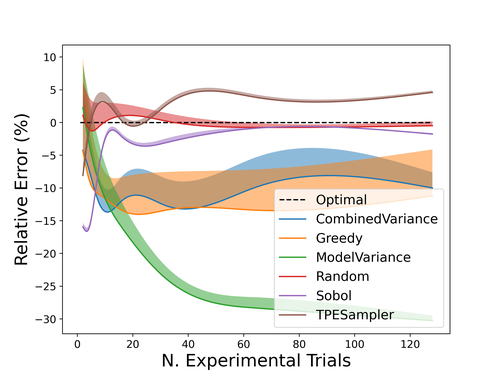}}
}
\end{minipage}
\begin{minipage}[t]{.29\linewidth}
   \hspace{-.8cm}\centering\raisebox{\dimexpr \topskip-\height-1.15cm}{%
\begin{tabular}{ll}
\hline 
  Bias (\%)& \\
\hline 
C. Var. & 48.90\\
Greedy & -1.38  \\
M. Var. &     -30.92 \\
Random       &    \textbf{5.79} \\
Sobol         &  8.06 \\
TPE        & 197.62 \\
\hline 
\end{tabular}
}
\end{minipage}
\caption{
Sampling bias in different hyperparameter search strategies on the NATS-Bench dataset. 
The relative error is calculated as the difference between the estimated performance of ResNets for given computational budget compared to the empirical performance of the method from an exhaustive search of topological variants. 
`Random' provides the an unbiased estimate closest to the true ResNets performance while remaining sample efficient.  
}
\label{fig:sampling_strategy}

\end{figure}
We apply each search strategy under identical experiment conditions for 10 repetitions with different random seeds and varying budgets (allocated trials) on the NATS-Bench benchmark \cite{dong2021nats} dataset. NATS-Bench includes a total of 39,419 experimental trials from an exhaustive search of topological variant networks evaluated on the CIFAR-10 dataset. We identify two topological variants: with Residual connections (`ResNets') \cite{resnets} and without. %

Of the strategies evaluated, those that accounted for variance underestimated the performance of `ResNets' as they sampled from regions of high variance and low average performance. In contrast, TPE overestimated the true model performance. Random and quasi-random approaches (Sobol) had the least biased estimates of the true performance. Counter-intuitively, the quasi-random approach performed poorly on a smaller budget, but eventually converged to perform similarly to random; \cref{fig:sampling_strategy}. 
Our results indicate that comparing methods optimized with different hyperparameter selection strategies will yield misleading results; however, we do not recommend the use of a specific strategy.

\section{Conclusions}
\label{sec:conclusions}

From the studies we analyzed unconsolidated experimental artifacts, which were the leading factor associated with unreproducible outcomes. The results indicate that failing to account for hyperparameter selection strategies as well as missing experimental artifacts can lead to incorrect analysis, since different search strategies and budgets can have significant effects on the statistical validity of the analysis. %

Missing artifacts and differences in hyperparameter selection are caused by poor experimental practices and can be mitigated by improved tooling. ML experiments may require large computation budgets of many experimental trials to produce a valid analysis\cite{dodge2019}. We observe that running multiple experimental trials required for the evaluation of a method using current best practices is cumbersome and error-prone. 

Current practices require researchers to use different tools for different parts of the experiment, such as algorithm design, distributed execution, and hyperparameter search. We conclude that reproducibility for ML experiments would be improved with a unified framework that facilitates large-scale experimentation and enforces the consolidation of experiment details and hyperparameter search values in concise artifacts.

\section*{Acknowledgement}

This work was supported by C-BRIC (one of six centers in JUMP, a
Semiconductor Research Corporation (SRC) program sponsored by DARPA), DARPA (HR00112190134) and the Army Research Office (W911NF2020053). The authors affirm that the views expressed herein are solely their own, and do not represent the views of the United States government or any agency thereof.

\bibliographystyle{IEEEtran}
\bibliography{main}

\begin{thebibliography}{1}
\providecommand{\url}[1]{#1}
\csname url@samestyle\endcsname
\providecommand{\newblock}{\relax}
\providecommand{\bibinfo}[2]{#2}
\providecommand{\BIBentrySTDinterwordspacing}{\spaceskip=0pt\relax}
\providecommand{\BIBentryALTinterwordstretchfactor}{4}
\providecommand{\BIBentryALTinterwordspacing}{\spaceskip=\fontdimen2\font plus
\BIBentryALTinterwordstretchfactor\fontdimen3\font minus
  \fontdimen4\font\relax}
\providecommand{\BIBforeignlanguage}[2]{{%
\expandafter\ifx\csname l@#1\endcsname\relax
\typeout{** WARNING: IEEEtran.bst: No hyphenation pattern has been}%
\typeout{** loaded for the language `#1'. Using the pattern for}%
\typeout{** the default language instead.}%
\else
\language=\csname l@#1\endcsname
\fi
#2}}
\providecommand{\BIBdecl}{\relax}
\BIBdecl

\bibitem{natureTransparentAI}
B.~Haibe-Kains, G.~A. Adam, A.~Hosny, F.~Khodakarami, {Massive Analysis Quality
  Control (MAQC) Society Board of Directors}, L.~Waldron, B.~Wang, C.~McIntosh,
  A.~Goldenberg, A.~Kundaje, C.~S. Greene, T.~Broderick, M.~M. Hoffman, J.~T.
  Leek, K.~Korthauer, W.~Huber, A.~Brazma, J.~Pineau, R.~Tibshirani, T.~Hastie,
  J.~P.~A. Ioannidis, J.~Quackenbush, and H.~J. W.~L. Aerts,
  ``\BIBforeignlanguage{en}{Transparency and reproducibility in artificial
  intelligence},'' \emph{\BIBforeignlanguage{en}{Nature}}, vol. 586, no. 7829,
  pp. E14--E16, Oct. 2020.

\bibitem{Shwartz-Ziv2021}
\BIBentryALTinterwordspacing
R.~Shwartz{-}Ziv and A.~Armon, ``Tabular data: Deep learning is not all you
  need,'' \emph{CoRR}, vol. abs/2106.03253, 2021. [Online]. Available:
  \url{https://arxiv.org/abs/2106.03253}
\BIBentrySTDinterwordspacing

\bibitem{mlDependencyUsageLibraryUsage}
J.~Han, S.~Deng, D.~Lo, C.~Zhi, J.~Yin, and X.~Xia, ``An empirical study of the
  dependency networks of deep learning libraries,'' in \emph{2020 IEEE
  International Conference on Software Maintenance and Evolution (ICSME)},
  2020, pp. 868--878.

\bibitem{emukit}
A.~Paleyes, M.~Pullin, M.~Mahsereci, N.~Lawrence, and J.~González, ``Emulation
  of physical processes with emukit,'' in \emph{Second Workshop on Machine
  Learning and the Physical Sciences, NeurIPS}, 2019.

\bibitem{Optuna}
\BIBentryALTinterwordspacing
T.~Akiba, S.~Sano, T.~Yanase, T.~Ohta, and M.~Koyama, ``Optuna: {A}
  next-generation hyperparameter optimization framework,'' \emph{CoRR}, vol.
  abs/1907.10902, 2019. [Online]. Available:
  \url{http://arxiv.org/abs/1907.10902}
\BIBentrySTDinterwordspacing

\bibitem{dong2021nats}
X.~Dong, L.~Liu, K.~Musial, and B.~Gabrys, ``{NATS-Bench}: Benchmarking nas
  algorithms for architecture topology and size,'' \emph{IEEE Transactions on
  Pattern Analysis and Machine Intelligence (TPAMI)}, 2021,
  \mbox{doi}:\url{10.1109/TPAMI.2021.3054824}.

\bibitem{resnets}
\BIBentryALTinterwordspacing
K.~He, X.~Zhang, S.~Ren, and J.~Sun, ``Deep residual learning for image
  recognition,'' \emph{CoRR}, vol. abs/1512.03385, 2015. [Online]. Available:
  \url{http://arxiv.org/abs/1512.03385}
\BIBentrySTDinterwordspacing

\bibitem{dodge2019}
\BIBentryALTinterwordspacing
J.~Dodge, S.~Gururangan, D.~Card, R.~Schwartz, and N.~A. Smith, ``Show your
  work: Improved reporting of experimental results,'' \emph{CoRR}, vol.
  abs/1909.03004, 2019. [Online]. Available:
  \url{http://arxiv.org/abs/1909.03004}
\BIBentrySTDinterwordspacing

\end{thebibliography}

\end{document}